\documentclass{ieeeaccess}
\usepackage{cite}
\usepackage{amsmath,amssymb,amsfonts}
\usepackage{algorithm,algorithmic}
\usepackage{graphicx}
\usepackage{textcomp}
\usepackage{xcolor}
\usepackage{color,soul}

\def\BibTeX{{\rm B\kern-.05em{\sc i\kern-.025em b}\kern-.08em
    T\kern-.1667em\lower.7ex\hbox{E}\kern-.125emX}}
\begin{document}
\history{Date of publication xxxx 00, 0000, date of current version xxxx 00, 0000.}
\doi{10.1109/ACCESS.2023.0322000}

\title{Unsupervised Work Behavior Pattern Extraction \\Based on Hierarchical Probabilistic Model}
\author{\uppercase{Issei Saito}\authorrefmark{1}, \uppercase{Tomoaki Nakamura}\authorrefmark{1},
\uppercase{Toshiyuki Hatta}\authorrefmark{2}, \uppercase{Wataru Fujita}\authorrefmark{2}, \uppercase{Shintaro Watanabe}\authorrefmark{2} and \uppercase{Shotaro Miwa}\authorrefmark{3}}

\address[1]{The University of Electro-Communications, Chofu, Tokyo, Japan }
\address[2]{Advanced Technology R\&D Center, Mitsubishi Electric Corporation }
\address[3]{Information Technology R\&D Center, Mitsubishi Electric Corporation }

\markboth
{Author \headeretal: Preparation of Papers for IEEE TRANSACTIONS and JOURNALS}
{Author \headeretal: Preparation of Papers for IEEE TRANSACTIONS and JOURNALS}

\corresp{Corresponding author: Issei Saito (e-mail:  i\_saito@radish.ee.uec.ac.jp).}

\begin{abstract}
Evolving consumer demands and market trends have led to businesses increasingly embracing a production approach that prioritizes flexibility and customization. 
Consequently, factory workers must engage in tasks that are more complex than before.
Thus, productivity depends on each worker's skills in assembling products. Therefore, analyzing the behavior of a worker is crucial for work improvement. 
However, manual analysis is time consuming and does not provide quick and accurate feedback.
Machine learning have been attempted to automate the analyses; however, most of these methods need several labels for training. 
To this end, we extend the Gaussian process hidden semi-Markov model (GP-HSMM), to enable the rapid and automated analysis of worker behavior without pre-training. The model does not require labeled data and can automatically and accurately segment continuous motions into motion classes.
The proposed model is a probabilistic model that hierarchically connects GP-HSMM and HSMM, enabling the extraction of behavioral patterns with different granularities.  
Furthermore, it mutually infers the parameters between the GP-HSMM and HSMM, resulting in accurate motion pattern extraction. 
We applied the proposed method to motion data in which workers assembled products at an actual production site.
The accuracy of behavior pattern extraction was evaluated using normalized Levenshtein distance (NLD). 
The smaller the value of NLD, the more accurate is the pattern extraction. 
The NLD of motion patterns captured by GP-HSMM and HSMM layers in our proposed method was 0.50 and 0.33, respectively, which are the smallest compared to that of the baseline methods. 

\end{abstract}

\begin{keywords}
behavior analysis, Gaussian process, hidden semi-Markov model, probabilistic generative model, unsupervised segmentation. 
\end{keywords}

\renewcommand{\boldsymbol}{\pmb}
\def\bq{\begin{equation}}
\def\eq{\end{equation}}
\def\beq{\begin{eqnarray}}
\def\eeq{\end{eqnarray}}
\def\ba{\begin{array}}
\def\ea{\end{array}}
\def\bc{\begin{center}}
\def\ec{\end{center}}

\def\dsum{\sum\limits}
\def\disp{\displaystyle}
\def\ejw{e^{j\omega}}
\def\ejwi{e^{j\omega_{i}}}
\def\e-jwi{e^{-j\omega_{i}}}
\def\dfrac#1#2{\disp{\frac{#1}{#2}}}
\def\teigi{\stackrel{\triangle}{=}}

\makeatletter
\def\lddots{\mathinner{\mkern1mu\raise\p@\vbox{\kern7\p@\hbox{.}}\mkern2mu
\raise4\p@\hbox{.}\mkern2mu\raise7\p@\hbox{.}\mkern1mu}}
\makeatother

\def\argmax{\mathop{\rm argmax}}

\def\baa{{ \boldsymbol a}}
\def\bb{{ \boldsymbol b}}
\def\bcc{{ \boldsymbol c}}
\def\bd{{ \boldsymbol d}}
\def\be{{ \boldsymbol e}}
\def\boldsymbolf{{ \boldsymbol f}}
\def\bg{{ \boldsymbol g}}
\def\bh{{ \boldsymbol h}}
\def\bi{{ \boldsymbol i}}
\def\bj{{ \boldsymbol j}}
\def\bk{{ \boldsymbol k}}
\def\bl{{ \boldsymbol l}}
\def\bm{{ \boldsymbol m}}
\def\bn{{ \boldsymbol n}}
\def\bo{{ \boldsymbol o}}
\def\bp{{ \boldsymbol p}}
\def\bqq{{ \boldsymbol q}}
\def\br{{ \boldsymbol r}}
\def\bs{{ \boldsymbol s}}
\def\bt{{ \boldsymbol t}}
\def\bu{{ \boldsymbol u}}
\def\bv{{ \boldsymbol v}}
\def\bw{{ \boldsymbol w}}
\def\bx{{ \boldsymbol x}}
\def\by{{ \boldsymbol y}}
\def\bz{{ \boldsymbol z}}

\def\bA{{ \boldsymbol A}}
\def\bB{{ \boldsymbol B}}
\def\bC{{ \boldsymbol C}}
\def\bD{{ \boldsymbol D}}
\def\bE{{ \boldsymbol E}}
\def\bF{{ \boldsymbol F}}
\def\bG{{ \boldsymbol G}}
\def\bH{{ \boldsymbol H}}
\def\bI{{ \boldsymbol I}}
\def\bJ{{ \boldsymbol J}}
\def\bK{{ \boldsymbol K}}
\def\bL{{ \boldsymbol L}}
\def\bM{{ \boldsymbol M}}
\def\bN{{ \boldsymbol N}}
\def\bO{{ \boldsymbol O}}
\def\bP{{ \boldsymbol P}}
\def\bQ{{ \boldsymbol Q}}
\def\bR{{ \boldsymbol R}}
\def\bS{{ \boldsymbol S}}
\def\bT{{ \boldsymbol T}}
\def\bU{{ \boldsymbol U}}
\def\bV{{ \boldsymbol V}}
\def\bW{{ \boldsymbol W}}
\def\bX{{ \boldsymbol X}}
\def\bY{{ \boldsymbol Y}}
\def\bZ{{ \boldsymbol Z}}

\def\b0{{\boldsymbol 0}}
\def\bPhi{{\boldsymbol\Phi}}
\def\bomega{{\boldsymbol\omega}}
\def\bLambda{{\boldsymbol\Lambda}}
\def\blambda{{\boldsymbol\lambda}}
\def\bmu{{\boldsymbol\mu}}
\def\bsigma{{\boldsymbol\sigma}}
\def\bnu{{\boldsymbol\nu}}
\def\bepsilon{{\boldsymbol\epsilon}}
\def\bphi{{\boldsymbol\phi}}
\def\bSigma{{\boldsymbol\Sigma}}
\def\bPhi{{\boldsymbol\Phi}}
\def\balpha{{\boldsymbol\alpha}}
\def\bTheta{{\boldsymbol\Theta}}
\def\btheta{{\boldsymbol\theta}}
\def\bGamma{{\boldsymbol\Gamma}}
\def\bPsi{{\boldsymbol\Psi}}
\def\bDelta{{\boldsymbol\Delta}}
\def\bPi{{\boldsymbol\Pi}}
\def\bXi{{\boldsymbol\Xi}}
\def\bxi{{\boldsymbol\xi}}
\def\bomega{{\boldsymbol\omega}}
\def\bOmega{{\boldsymbol\Omega}}

\def\balpha{{\boldsymbol\alpha}}
\def\bbeta{{\boldsymbol\beta}}

\titlepgskip=-21pt

\maketitle
\section{Introduction}

\PARstart{A}{nalyzing} human movement in industrial work environments is significant because of its implications for safety, efficiency, and productivity. 
Such an analysis facilitates understanding how workers interact with their environment, machinery, and tools, with the goal of optimizing work processes, reducing the risk of injury, and enhancing the workplace environment. 

Conventionally, industries employ line production to produce standardized products in large quantities. 
Recently, products have been customized to meet diverse consumer needs, leading to the production of a wide range of products in small quantities. 
Consequently, the assembly work has changed from simple to complex tasks involving multiple processes.
As work becomes more complex, the impact of individual productivity on overall productivity increases. 
Therefore, it is important to conduct work analyses to optimize the workflow.
To date, the VTR method, in which videos are recorded and analyzed, has been used for work analysis at production sites. 
Additionally, the stopwatch method\cite{Budiman_2019} has been used to manually identify time-consuming behaviors and incorrect procedures by measuring the time required for each elementary task using a stopwatch.
However, because these analyses are performed manually, they require considerable time and effort.
This results in an inability to quickly return the analysis results to the workers.
In addition, analysts’ heavy workloads cause errors in analysis.
To solve this problem, recent studies have been conducted to automatically analyze work using machine learning\cite{10.1145/3534572,informatics5020026,inproceedingsaa,inproceedings}.
In such studies, work analyses were realized through supervised learning. 
However, these methods require numerous labeled training data. Methods that use multiple labeled data are unsuitable for work analysis in high-mix, low-volume production for two reasons.

\begin{enumerate}
   \item It is difficult to apply a model trained on labeled data from one worker to the analysis of others owing to variations in how people perform the same task.
   Therefore, training data must be taken for each worker, which requires a considerable amount of data.
   \item In real workplaces, products are frequently changed to meet changing customer needs, and the work changes each time. 
   Therefore, new training data have to be obtained frequently to cope with these changes.
\end{enumerate}

\begin{figure}[t]
	\begin{center}
	\includegraphics[scale=0.5]{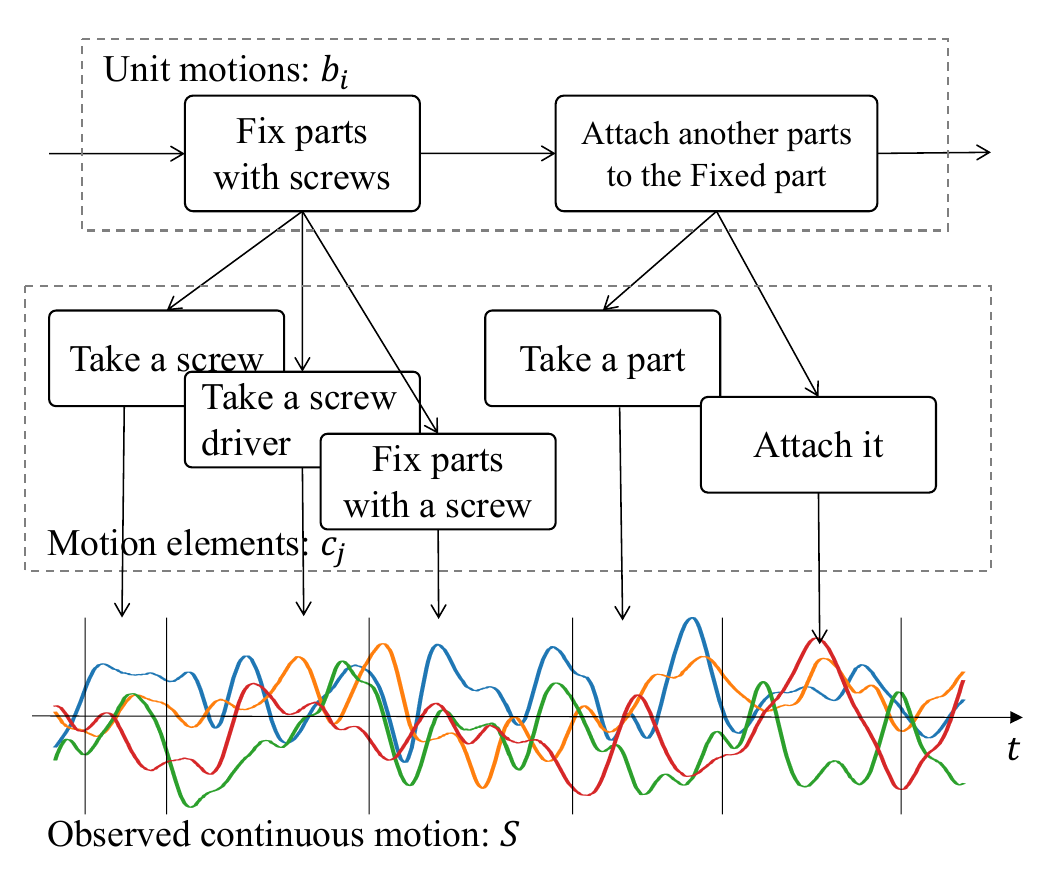}
	\caption{Overview}
	\label{fig:overview}
	\end{center}
	\vspace{-0.4cm}
\end{figure}

For analyzing workers' behavior, rapidly processing data through automated means is desirable. 
However, the practical implementation of this approach is challenging owing to the limitations of supervised methods, as previously discussed, and the underutilization of unsupervised approaches in industrial analysis. 
To bridge this gap,  a swift and accurate methodology employing unsupervised models without pre-training is required.
In this context, we propose the use of Gaussian process-hidden semi-Markov model (GP-HSMM)\cite{nakamura2017segmenting} as an unsupervised human behavior analysis method that does not require label data, and can segment even complex behaviors with high accuracy. 
This method is a probabilistic generative model (PGM) that estimates segments from skeletal coordinate time-series data using a Gaussian process and HSMM.
Although it is more accurate than conventional hidden Markov model (HMM)-based methods \cite{fox2011joint, matsubara2014autoplait},
its use has been limited to experiments using motion capture data and has not yet progressed to real-world applications. By applying segmentation to real-world data, the automatic discretization of continuous data may become feasible, enabling practical applications in industrial work analysis.
In this study, we extend the GP-HSMM to propose a hierarchical model that can segment actions as well as tasks composed of combinations of actions.
Here, the smallest action unit is called a ``motion element,” and a task composed of combinations of them is called a ``unit motion.” 
For example, ``picking up a screw with the right hand,'' ``holding a screwdriver with the opposite hand,'' ``inserting a screw into a screw hole of a part,'' and ``putting down a screwdriver,'' each of which is a motion element, are combined into the unit motion ``installing a screw with a screwdriver.” 
In particular, work can be analyzed at different granularities by further segmenting the motion elements obtained by segmenting the time-series data and determining their cohesion.
In this study, we propose a hierarchical PGM capable of performing unsupervised segmentation of motion into motion elements and unit motions, which are meaningful collections (Figure\ref{fig:overview}).

The simplest way to implement such a two-layer model is to employ a GP-HSMM to segment the continuous skeletal coordinates into motion elements in the lower layer, and then segment the discretized class sequence in the upper layer using the word segmentation method. 
However, this method has two limitations.
First, if there is an error in the estimation of the motion element using the GP-HSMM, the motion elements with errors are directly segmented in the upper layer.
This reduces the accuracy of the segmentation estimation of the unit motion.
To solve this problem, the proposed model introduces hierarchical mutual learning to improve the segmentation accuracy of the GP-HSMM.
The information of the motion elements composing the unit motion is used in the lower layer (GP-HSMM) learning to reduce segmentation errors.
Second, the same-role unit motion can be classified into different unit motion classes using a simple word-segmentation method. 
This occurs when the sequences of motion elements differ owing to fluctuations in the action or individual differences.
From a task analysis perspective, it is desirable to classify behaviors with the same meaning into the same class, even if they are composed of slightly different sequences of motion elements. 
The proposed method addresses this problem by introducing a probability distribution to generate the elements of each unit action.

Two types of experiments were conducted using the 6-dimensional time-series data of both wrists of three workers.
In Experiment 1, to verify whether the proposed method can solve the first problem, we show that the segmentation accuracy of the proposed method is better than that of GP-HSMM without mutual learning. 
In Experiment 2, to verify whether the proposed method can solve the second problem, we change the probability distribution for generating the unit motion and test its effect on the segmentation result of the unit motion.

The main contributions of this study are as follows: 
\begin{itemize}
\item A novel two-layer PGM based on GP-HSMM for work behavior segmentation is proposed.
\item An algorithm to infer the parameters of each layer mutually, enhancing segmentation accuracy, is proposed.
\item The proposed method achieves higher accuracy than the baseline method when applied to the real motion data of the cell production operation. 
\end{itemize}

\section{Related work}
\label{sec:related}
Supervised learning methods can now accurately capture patterns in time-series data and analyze human behavior\cite{lea2017temporal}\cite{9067967}\cite{lea2017temporal}\cite{yeung2016end}\cite{diba2018spatio}. 
However, these methods require pre-training using numerous labeled data. 
Therefore, it is difficult for these methods to be applied to analyze real work involving many types of processes and requiring rapid feedback. 

Subsequently, half-supervised learning segmentation analysis methods, which do not require multiple labeled data, were proposed by \cite{bojanowski2014weakly}\cite{huang2016connectionist}\cite{richard2017weakly}.
These methods require fewer training data; however, they must be performed beforehand. 
These methods can be applied to analyze work using predetermined procedures.
However, in practice, procedures can change; in this case, methods that use half-supervised learning cannot be utilized.

To overcome this limitation, \cite{gmmunsupervised,fox2011joint,matsubara2014autoplait} proposed human motion analysis methods using unsupervised learning. 
These methods do not require pre-learning.
\cite{gmmunsupervised} used clustering with the Gaussian mixture model (GMM) and demonstrated that their method could segment a movie accurately without pre-training.
This study assumed that the workers performed the same motion only once during a task.
Therefore, the segmentation accuracy decreased when the worker performed repetitive motions in the data. 
In a practical industrial production scenario, there are repetitive motions, such as screwing multiple places to fix the parts.
Hence, it is difficult to apply this method to analyze real work.
\cite{fox2011joint,matsubara2014autoplait} proposed models that use an HMM to infer the segments stochastically.
Fox et al. proposed a method using HMMs for unsupervised segmentation of time-series skeletal information obtained from motion capture data\cite{fox2011joint}. 
This method extracts continuous data points that are classified into the same class as segments.
However, HMMs often produce shorter segments because states tend to transfer to other states in the short term. 
Furthermore, Matsubara et al. proposed the segmentation method AutoPlait, which uses multiple HMMs, each of which represents a type of motion pattern \cite{matsubara2014autoplait}. 
This approach segments time-series data when the HMM switches to another. 
However, HMMs use the mean and standard deviation to represent time-series data, which is considerably too simple to represent complex sequences, such as motion.

To overcome this limitation, we propose GP-HSMM\cite{ nakamura2017segmenting}, which represents motion trajectories using Gaussian processes and models the duration of motion using HSMM \cite{yu2010hidden}.
This method can segment complex motion sequences more accurately than existing methods.
Therefore, in this study, we propose a hierarchical segmentation method based on GP-HSMM, which further segments a sequence segmented by GP-HSMM. 
Moreover, we propose mutual learning between hierarchies in the models to improve segmentation accuracy. 

Studies have been conducted on the learning of such hierarchical motion structures. 
\cite{7251510} proposed a two-level segmentation method to accurately capture more complex human motions by decomposing motions into motion primitives. 
However, this method performs segmentation based on the contact relationships between objects, and then performs segmentation using motion–property heuristics. 
Therefore, this method can only be applied to a limited number of situations. 
Taniguchi et al. proposed a method to learn elementary and unit motions, which are segments of elementary motion, from the joint angles of the upper body \cite{article, taniguchi2011unsupervised}. 
In a study on fish behavior analysis, a Gaussian mixture model (GMM) was used to estimate the unit motion of fish from symbolic action sequences \cite{10.1371/journal.pcbi.1009672}. 
However, in these methods, each of the two layers learns independently, and errors in the lower layer directly cause errors in the upper layer.

In the field of natural language processing, studies have been conducted on the unsupervised segmentation of sentences. 
For example, an unsupervised morphological analysis method was proposed to segment sentences into words (\cite{gold, mochihashi-etal-2009-bayesian,uchiumi-etal-2015-inducing}). 
Goldwater et al. \ proposed a method for segmenting sentences into words by estimating the parameters of a bigram language model based on hierarchical Dirichlet processes \cite{gold}. 
Mochihashi et al. \ proposed a method for word segmentation that uses an n-gram language model based on the hierarchical Pitman--Yor process \cite{mochihashi-etal-2009-bayesian}. 
Uchiumi et al. extended NPYLM to a Pitman--Yor hidden semi-Markov model (PY-HSMM) and realized segmenting sentences into words as well as estimating part of speech of words  \cite{uchiumi-etal-2015-inducing}.

Additionally, there are studies using unsupervised learning methods for behavior analysis \cite{khanfar2022application,un2}. 
Khanfar et al. applied unsupervised machine learning to classify driver behavior in work zones in Qatar, providing patterns to improve road safety and traffic management in these areas \cite{khanfar2022application}. 
Wang et al. applied similarity graphs to the clickstream of an online service and made it possible to extract previously unknown behaviors \cite{un2}.
Although these studies applied unsupervised learning to analyze human behavior, they did not use human movements. 
To analyze worker behaviors in various industrial fields, it is desirable to analyze human movements without using information obtained from specific devices.
Some studies have applied unsupervised learning to human movements \cite{walk,rnn,li}; however, they focused on clustering or recognition of human actions and have not been applied to behavior analysis in the industrial field.
\section{Proposed model}
\label{sec: proposed model}
\subsection{Generative process}
\label{sec: generative process}
\begin{figure}[t]
	\begin{center}
	\includegraphics[scale=0.7]{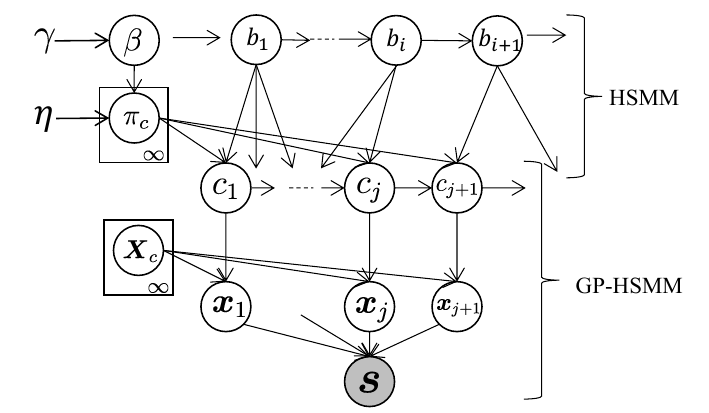}
	\caption{ Graphical model of GP-HSMM-BA }
	\label{fig:model}
	\end{center}
	\vspace{-0.4cm}
\end{figure}

Figure \ref{fig:model} shows the proposed graphical model, which is a PGM, in which the bottom layer is GP-HSMM, and the upper layer is HSMM.
This assumes the following generative process and generates a series of motions $S$:
The unit motion class $b_{i}(i=1,2,...)$ is determined by the previous unit motion $b_{i-1}$:
\begin{eqnarray}
b_i &\sim& P(b|b_{i-1}). 
\end{eqnarray}
%
The motion element class $ c_j $ is determined by the previous class $ c_{j-1} $, corresponding unit motion $b_i$, and transition probability $\pi_c$:
\begin{eqnarray}
c_j &\sim& P(c|c_{j-1}, \pi_c, b_i). 
\end{eqnarray}
%
In this process, it is assumed that a series of motion elements (e.g. $c_{j-1}, c_j, c_{j+1}, \cdots$) are generated from a single unit motion $b_i$. 
Segment $\bx_j$ corresponding to the motion element class $c_j$ is generated by a Gaussian process with parameter $\bX_{c_j}$:
\begin{eqnarray}
\bx_j &\sim& {\mathcal GP}(\bx|\bX_{c_j}).
\end{eqnarray}
%
$\bx_j$ is a time series composed of multiple data points and, therefore, this process generates a series of data points in the observation from the single motion element class $c_j$.
The observed motion sequence $\bS$ is generated by concatenating $\bx$.

The observed motion sequences can be divided and classified into short-term motion elements $c_j$ and long-term unit motions $b_i$ by estimating model parameters in an unsupervised manner. 
As explained earlier, the generative process assumes that the series of motion elements and data points in the observation are generated from single classes $b_i$ and $c_j$, respectively. 
That is, the length of the classified data in each class is also estimated during the inference process.
This is not an HMM where a single data point is classified into a single class, but HSMM \cite{yu2010hidden}. 

\subsection{Gaussian Process}
The lower-layer GP-HSMM utilizes Gaussian processes to represent the continuous trajectory of the output $x_t$ at timestep $t$ in the motion element $\bx$.
In Gaussian processes, when the sets $(\bt, \bX)$ of output $x_t$ and timestep $t$ in the same motion element are obtained, the predictive distribution of the output $\hat{x}$ at timestep $\hat{t}$ becomes a Gaussian distribution:
\begin{equation}
\label{equ:gp_predict}
p(\hat{x} | \hat{t}, \bX,\bt) \propto {\mathcal N}(\bk^T \bC^{-1} \bX, k(\hat{t},\hat{t})-\bk^T \bC^{-1} \bk), 
\end{equation}
where $k(\cdot, \cdot )$ denotes the kernel function. $\bC$ is a matrix whose $p$ row and $q$ column elements, $C(t_p, t_q)$, are
\begin{equation}
\label{equ:covariance_func}
C(t_p, t_q)=k(t_p, t_q)+\phi^{-1} \delta_{pq} ,
\end{equation}
$\phi$ is a hyperparameter that represents the noise in the observation.
$\bk$ is a vector whose $p$-th element is $k(t_p,\hat{t})$.
In this study, the following kernel function was used:
\begin{equation}
\label{equ:kernel_default}
k(t_p,t_q)=\theta_{0}\exp(-\frac{1}{2}\theta_1 || t_p-t_q ||^{2}) + \theta_2 + \theta_3 t_p t_q. 
\end{equation}
$\theta_*$ is a hyperparameter of the kernel.

When the output is a multidimensional vector $\bx_t=(x_t^{(1)}, \cdots, x_t^{(d)},\cdots)$,
we assume that each dimensional output is generated independently. 
The probability $\mathcal GP(\bx|\bX_c)$ that the observed value $\bx$ at time $t$ is generated by a Gaussian process corresponding to class $c$ is computed as 
\begin{equation}
\label{equ:multi_dim}
{\mathcal GP}( \bx |\bX_c) = \prod_d^D p(x_t^{(d)} | t , \bX_c^{(d)} ). 
\end{equation}

\section{Parameter Inference}
\subsection{Inference Algorithm}
\label{sec: Inference Algorithm}
The proposed model is a double hierarchical model, which makes it difficult to infer parameters.
Thus, we apply a message-passing method proposed in the Serket framework\cite{serket_nakamura18,taniguchi2020neuro}.
The Serket framework enables the connection and mutual training of the GP-HSMM and HSMM; the parameters of each model are mutually inferred using Algorithm \ref{alg:gibbs}. 

\begin{algorithm}[t]
\small
\caption{Mutual parameter update}
\label{alg:gibbs}
\begin{algorithmic}[1]
\STATE// Initialization
\STATE Set $P(\bC|\bB)$ to uniform distribution

\STATE
\FOR{$m = 1$ to $M$}
	\STATE // Learning of lower layer
	\STATE $\bC \sim GP-HSMM(\bS, P(\bC|\bB)) $
	\STATE
	\STATE // Learning of higher layer
	\STATE $\bB \sim HSMM(\bC)$ 
	\STATE
	\STATE// Parameter update
	\STATE Update $P(\bC|\bB)$ from $\bB$ and $\bC$
\ENDFOR
\end{algorithmic}
\end{algorithm}
\begin{algorithm}[t]
\small

\caption{Forward filtering--backward sampling of GP-HSMM.}
\label{alg:ffbs}
\begin{algorithmic}[1]
\STATE // Forward filtering
\FOR{ $t=1$ to $T$}
	\FOR{ $k=1$ to $K$}
		\FOR{ $c=1$ to $C$}
			\STATE Compute $\alpha[t][k][c]$
		\ENDFOR 
	\ENDFOR  
\ENDFOR  
\STATE
\STATE // Backward sampling
\STATE $\bar{C}$ = [ ]
\STATE $\bar{X}$ = [ ]
\STATE $t = T$
\WHILE{ $t>0$ }
	\STATE $k, c \sim \alpha[t][k][c] P(c | c')$
	\STATE $\bx = \bs_{t-k:t}$
	\STATE $t = t-k$
	\STATE $\bar{C} = [ c, \bar{C}] $  ~~~~~ //$c$ is prepended to $\bar{C}$
	\STATE $\bar{X} = [ \bx, \bar{X} ]$  ~~~~~//$\bx$ is prepended to $\bar{X}$
\ENDWHILE

\STATE
\STATE return $\bar{C}, \bar{X}$
\end{algorithmic}
\end{algorithm}

First, in the lower layer, the observed motion waveform $\bS$ is segmented using GP-HSMM, and the motion element class sequence $\bC$ is sampled.
Subsequently, the obtained motion element class sequence is segmented using the HSMM in the upper layer, and the unit motion sequence $\bB$ is sampled.
The upper layer computes the conditional probability $P(\bC|\bB)$ to generate the motion element class $\bC$ from the segmented unit motion $\bB$ and sends it to the lower layer (GP-HSMM).
The GP-HSMM uses the received $P(\bC|\bB)$ as a prior distribution of the motion element and resamples the motion element classes.
This mutual update is repeated $M$ times to optimize the parameters.

GP-HSMM and HSMM use the forward filtering--backward sampling algorithm \cite{uchiumi-etal-2015-inducing} to efficiently sample segment lengths and classes using Algorithm \ref{alg:ffbs}.
Forward filtering in GP-HSMM computes the forward probability that a subsequence of length $k$ before timestep $t$ of the motion sequence becomes motion element class $c$ as follows: 
\begin{eqnarray}
\label{eq:forward_gp}
\alpha_m [t][k][c] =  {\mathcal GP}(\bx_{t-k:t} | \bX_c) P(c|b_i) P_{len}(k|\lambda_p)  \nonumber \\
~~~~~ \times \sum_{k'=1}^{K} \sum_{c'=1}^{C}P(c|c', \pi_{c'} )\alpha_m [t-k][k'][c'], 
\end{eqnarray}
where $P_{len}(k|\lambda_p)$ is a Poisson distribution with $\lambda_p$ being a parameter that determines the segment length. 
$K$ is the max length of  the segment, and $C$ is the number of motion element classes.
In addition, the product of the expert (PoE) approximation was used to calculate the transition probabilities, $P(c | c',\pi_{c'}, b_i ) \approx \propto P(c | c',\pi_{c'}) P(c|b_i)$. 
$P(c|b_i)$ is the probability that class $c$ of the motion element occurs from the unit motion computed in the upper layer (HSMM). 
This probability can be used to constrain the motion elements that comprise the unit motion to learn the motion elements. 
Class series $\bC$ is sampled from this forward probability.

Subsequently, in the upper layer, the unit motion is sampled by segmenting the motion element class sequence $\bC$.
Forward filtering calculates the probability that a subsequence of length $k$ before time step $j$ becomes a unit of motion $b$, as shown in the following equation:
\begin{eqnarray}
\label{eq:hs}
\alpha_b[j][k][b] =  P(c_{j-k:j}|b) P_{len}(k|\lambda_b)  \nonumber \\
~~~~~~~~ \times \sum_{k'=1}^{K'} \sum_{b'=1}^{B} P(b|b')\alpha_b [j-k][k'][b'], 
\end{eqnarray}
where $K'$ is the max length of  the segment, and $B$ is the number of unit motion classes.
The unit motion sequence $\bB$ is sampled from the forward probability.
$P(c|b_i)$ is updated from the sampled $\bB$ and $\bC$, and is used in the GP-HSMM calculation.

By repeating the above calculations in the following procedure, the lower and upper layers interact with each other to learn the motion elements and unit motions.
\begin{enumerate}
\item Sample motion element sequence $\bC$ from motion waveform $\bS$
\item Sample unit motion sequence $\bB$ from motion element class sequence $\bC$ 
\item Updating probability $P(c|b_i)$ in which motion elements are generated from each unit motion
\end{enumerate}
\subsection{Emission Probability of HSMM and Prior Probability of GP-HSMM}
\label{emit}
$P(c_{j-k:j}|b)$ in Eq. (\ref{eq:hs}) is the emission probability of the motion elements $c_{j-k:j}$ from the unit motion $b$ in HSMM, and  $P(c|b_i) $ in Eq. (\ref{eq:forward_gp}) is the prior probability of the motion element in GP-HSMM.
The formulation of these probabilities affects the segmentation performance. 
In this study, we considered the following three patterns:
\begin{itemize}
	\item {\bf Formulation with word segmentation:} This is the most straightforward way to use a similar idea as unsupervised word segmentation. The emission probability is computed using the unigram word segmentation model as follows: 
	\begin{eqnarray}
	P(c_{j-k:j}|b) &=& \frac{N_{c_{j-k:j}} + \alpha}{N_{\rm all} + \alpha V }. 
	\label{uu}
	\end{eqnarray}   
	$\alpha$ is the parameter of the Dirichlet prior distribution. 
	$N_{c_{j-k:j}}$ is the number of occurrences of unit motion consisting of exactly the same sequence $c_{j-k:j}$. 
	$N_{\rm all}$ is the total number of unit motions, and $V$ is the number of types of unit motions.

        ~~In this model, if elements in subsequences, even if it is only one element, are different, they are considered different unit motions.
	We call the HSMM using this emission probability word segmentation HSMM (WS HSMM) in this paper.

       ~~Because the unit motion is composed of a sequence of motion elements in exactly the same pattern, it cannot be categorized, and $P(c|b)$ is not computed in each unit motion class $b$.
      Therefore, the probability of generating the motion element $P(c|b)$, is computed from all the segmented motion elements according to the position of $c$ and used as a prior distribution in Eq. (\ref{eq:forward_gp}): 
	 \begin{align}
	 &P(c|b) \propto \nonumber \\
	 &~~ \begin{cases}
	 \mathrm{ count_{begin}}( c ) + \mu 	\\
	  	~~~ \text{: if position of $c$ is included in begin }\\ 
	 	~~~~ \text{ of unit motions in $(m-1)$-th inference} \\
	\mathrm{count_{trans}}(c, \bar{c}) + \mu \\
	  	~~~ \text{: if position of $c$ is the middle  }\\
	   	~~~~\text{of the unit motion in $(m-1)$-th inference}\\
	(\mathrm{ count_{trans}}(c, \bar{c})+\mu ) (\mathrm{count_{end}}( c) + \mu) 	\\
	        ~~~ \text{: if position of $c$ is included in end of   }\\ 
	        ~~~~\text{unit motions in $(m-1)$-th inference}\\
	  \end{cases} \nonumber \\ \label{eq:pcw}
	\end{align}
	$\mathrm{count_{begin}}( c )$ and $\mathrm{ count_{end}}( c )$ are the number of occurrences of the motion element $c$ at the beginning and end of the segmented unit motion, respectively.
	$\mathrm{count_{trans}}(c, \bar{c})$ is the number of occurrences of $c$ after the preceding $\bar{c}$.
	The $\mu$ is a parameter of the Dirichlet prior distribution.
	These probabilities are multinomial distributions representing the probability of occurrence of the motion element $c$ at the beginning, in the middle after $\bar{c}$, and at the end of the unit motion, respectively.

~~\item {\bf Formulation with motion element unigram:}
This is a model in which motion element $c$ is generated from unit motion class $b$ independently, and the probability is expressed as follows: 
	\begin{eqnarray}
	P(c_{j-k:j}|b) &=& \prod_{t=j-k}^{j}\frac{N_{b,c_t} + \alpha}{N_{b} +  \alpha C}, 
	\label{bu}
	\end{eqnarray}
	$N_{b}$ is the total number of motion elements classified into class $b$, and $N_{b,c_t}$ is the number of element motions whose class is $c_t$ among them.
	We call HSMM with this emission probability motion element unigram HSMM (ME-U HSMM).

      ~~Unit motions are composed of multiple similar motion element patterns; therefore, the probability that motion element $c$ is generated according to its position in unit motion $b$ can be computed as follows: 
      \begin{align}
	  \label{prior}
	  &P(c|b) \propto  \nonumber \\
	  &~~\begin{cases}
	  \mathrm{ count_{begin}}( c,b ) + \mu 	\\
	  ~~~ \text{: if position of $c$ is included in begin of }\\
	  ~~~~ \text{the unit motion in $(m-1)$-th inference}\\
	  \mathrm{ count_{trans}}(  \bar{c},c,b) + \mu \\
	  ~~~ \text{: if position of $c$ is the middle of }\\
	  ~~~~ \text{the unit motion in $(m-1)$-th inference}\\
	  (\mathrm{ count_{trans}}(  \bar{c},c ,b)+\mu )  (\mathrm{ count_{end}}(c,b) + \mu) 	\\
	  ~~~ \text{: if position of $c$ is included in end of  }\\
	  ~~~~ \text{the unit motion in $(m-1)$-th inference}\\
	  \end{cases} 
	  \end{align}
	$\mathrm{count_{begin}}( c, b )$ and $\mathrm{ count_{end}}( c, b )$ are the number of times the motion element $c$ occurs at the beginning and end of the segmented motion elements classified into unit motion $b$, respectively.
	${\mathrm{count_{trans}}(c, \bar{c}}, b)$ is the number of times $c$ occurs after one previous motion element $\bar{c}$ in the motion elements classified into unit motion $b$.
	$\mu$ is a parameter of the Dirichlet prior distribution.
	These probabilities are multinomial distributions representing the probability of occurrence of the motion element $c$ in the unit motion $b$ at the beginning, the probability of occurrence of the motion element $c$ after $\bar{c}$ in the middle, and the probability of occurrence at the end of the unit motion $b$.
      \item {\bf Formulation with motion element bigram:} In the motion element unigram model, the motion elements are independent of their order; the motion elements are generated independently for each unit motion. 
      Conversely, the motion element bigram model uses bigrams to represent the order of the motion elements in the unit motions and is expressed as follows:
	\begin{eqnarray}
	P(c_{j-k:j}|b) &=& \prod_{t=j-k+1}^{j}\frac{N_{z,c_{t-1},c_t} + \alpha}{N_{z,c_{t-1}} + \alpha C }, 
	\label{bb}
	\end{eqnarray}
	$N_{b,c_{t-1},c_t}$ is the number of transitions from $c_{t-1}$ to $c_{t}$ in the motion elements classified into a unit motion $b$, and $N_{b,c_{t-1}}$ is the number of times the motion element $c_{t-1}$ occurs in the motion elements classified into unit motion $b$. 
	We call HSMM with this emission probability motion element bigram HSMM (ME-B HSMM).
	The probability of generating a motion element for each class is the same as in equation (\ref{prior}).
	\end{itemize}

\section{Experiment}
The proposed method was validated by using it to segment the motion data of cell production operation.
\subsection{Experimental Setup}
\begin{figure}[t]
	\begin{center}
	\includegraphics[scale=0.3]{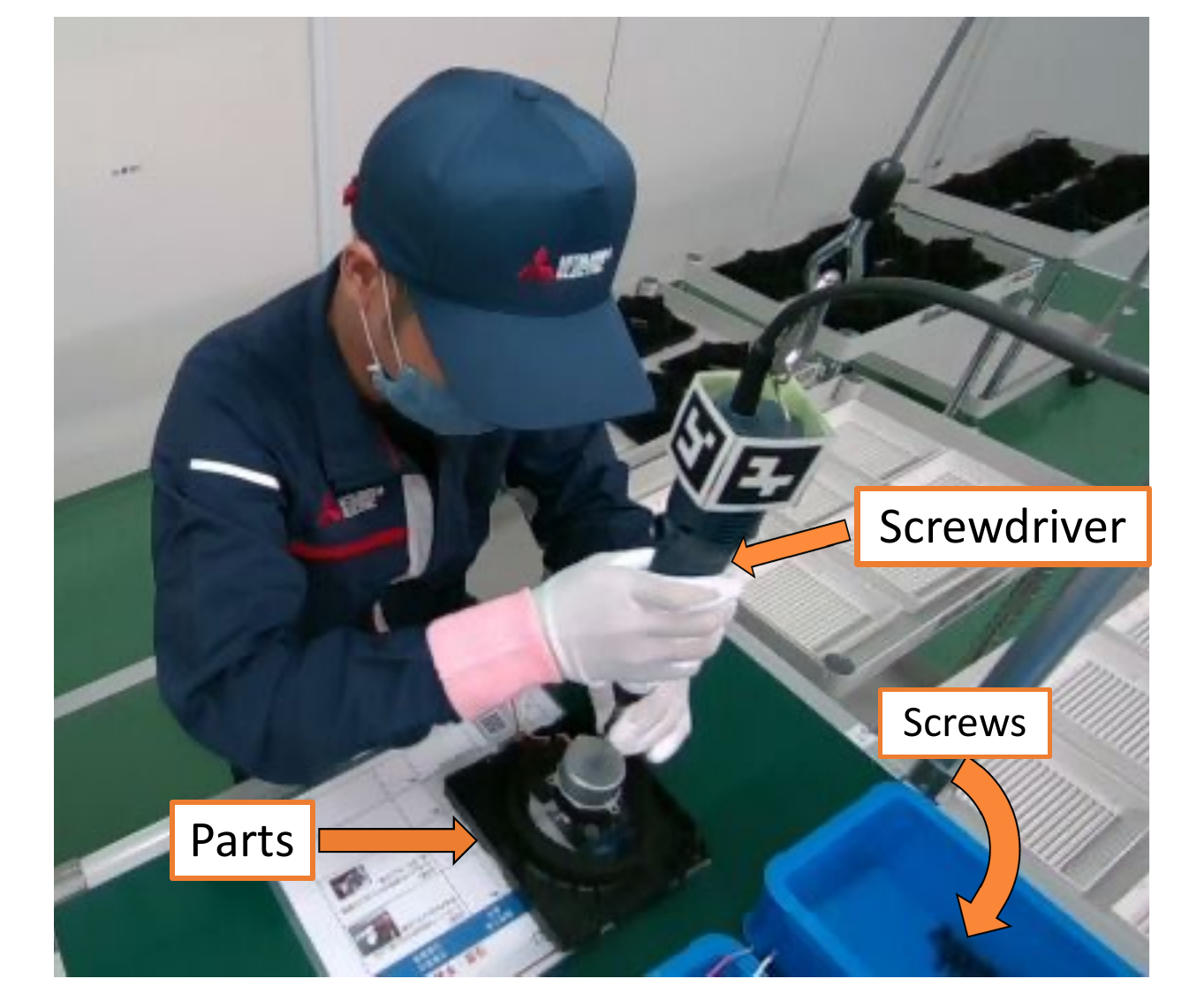}
	\caption{ Work scenario }
	\label{fig:senerary}
	\end{center}
	\vspace{-0.4cm}
\end{figure}

The six-dimensional time-series positions of the left and right wrists of three workers engaged in fan assembly were used.
The workers wore a pink wristband on the right wrist and a red wristband on the left wrist. 
The coordinates of the wrists were obtained by tracking their colors in the recorded RGB-D data. 
To mitigate occlusion, the coordinates obtained from three RGB-D cameras placed in different positions were utilized.
Each worker repeated the procedure in Table \ref{tbl:class} 36 times, as shown in Figure \ref{fig:senerary}.
The workers were novices at assembling the products.
Therefore, we used 108 motion sequences whose length ranges from 29 to 65 s, composed of five frames per second. 

We empirically set hyper parameters in Eq. (\ref{equ:kernel_default}) to $\theta_{0} = 1$, $\theta_{1 }= 1$, $\theta_{2 }= 0$, $\theta_{3}= 16$.
These are the same values used in our previous study
\cite{10.3389/frobt.2019.00115}.
The number of classes of motion elements and unit motions was set to $C=12$ and $B=8$, respectively, and other hyperparameters were set to $\alpha=10, \mu=0.1$.

\begin{table}[t]
\small
\caption{Task procedures}
\label{tbl:class}
\centering
\begin{tabular}{ccl}\hline
 Procedure & Motion  &  Description\\ 
                 &  Label    &                   \\ \hline
1 & 1 &Take part A from the cart and \\
   &  & ~~~~place it on the workspace\\
2 & 2 & Take part B from the box\\
3 & 3 & Attach part B to part A\\
4 & 4 & Take a screw from the box\\
5 & 5 & Fix part B and part A with the screw\\
6 & 4 & Take a screw from the box\\
7 & 5 & Fix part B and part A with the screw\\
8 & 4 & Take a screw from the box\\
9 & 5 & Fix part B and part A with the screw\\
10 & 6 & Take part C from the box\\
     &   & ~~~~and attach it to part B\\
11 & 7 & Take part D from the cart\\
12 & 8 & Attach part D to part A and\\
     &    & ~~place the finished product on the cart\\
\hline
\end{tabular}
\end{table}

Segmentation was performed using the following four methods, and each was trained for 30 iterations:
\begin{itemize}
   \item {\bf GP-HSMM:} A method for segmenting time-series skeletal coordinates into motion elements using GP-HSMM alone.
   \item {\bf GP-HSMM+WS HSMM:} A method for segmenting time-series skeletal coordinates with GP-HSMM, and then segmenting the segmented sequence of motion elements with the simple word segmentation (WS) HSMM.
   For learning, we used mutual learning as described in Section \ref{sec: Inference Algorithm}.
   \item {\bf GP-HSMM+ME-U HSMM:} A method for segmenting time-series skeletal coordinates with GP-HSMM, and then segmenting the segmented sequence of motion elements with the motion element unigram HSMM.
   For learning, we employed mutual learning as described in Section \ref{sec: Inference Algorithm}.

   \item {\bf GP-HSMM+ME-B HSMM:} A method for segmenting time-series skeletal coordinates with GP-HSMM, and then segmenting the segmented sequence of motion elements with the motion element bigram HSMM.
   For learning, we used mutual learning as described in Section \ref{sec: Inference Algorithm}.
\end{itemize}

The normalized Levenshtein distance (NLD) between the segmented series $\bC$ and the correct label series $\hat{\bC}$ in the following equation was used as an evaluation index:
\begin{eqnarray}
\bar{d}(\bC,\hat{\bC}) = \frac{d(\bC,\hat{\bC})}{ \max(|\bC|, |\hat{\bC}|)}, 
\end{eqnarray}
$d(\bC,\bar{\bC})$ is the Levenshtein distance \cite{levenshtein1966binary} between the two series and $|\bar{\bC}|$ is the length of the series. The NLD assumes values between zero and one. The closer it is to the correct labels, the closer it is to zero.
The unit motion $\bB$ was evaluated in the same manner as the motion elements $\bC$.
Because there is an initial value dependence in learning, each method was segmented ten times with different initial values, and the result with the maximum likelihood was used for evaluation.

\subsection{Segmentatation of motion elements}
\begin{figure}[t]
	\begin{center}
	\includegraphics[scale=0.35]{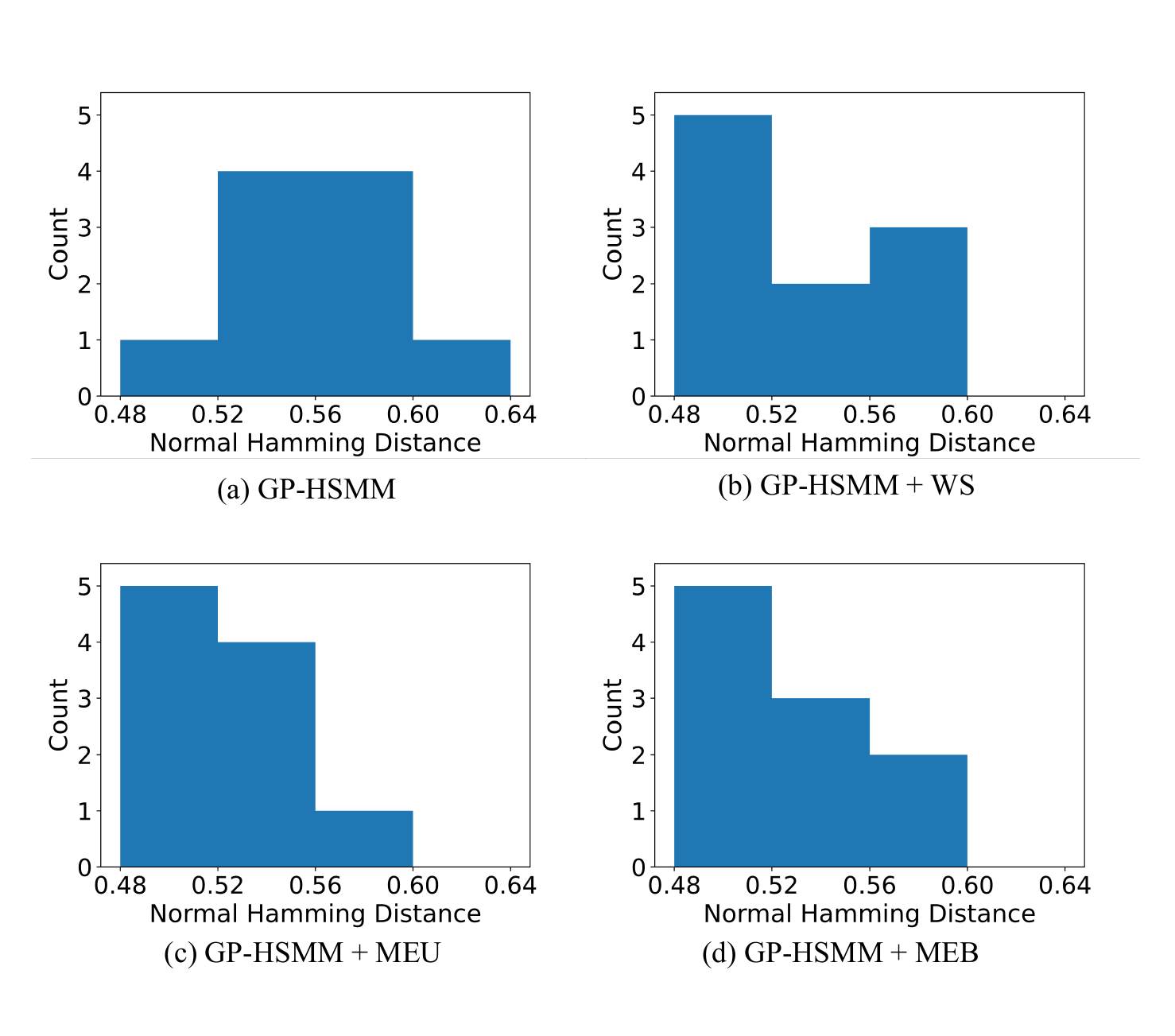}
	\vspace{-1.25cm}
	\caption{ Normalized Levenshtein distance of 10 trials }
	\label{fig:ham}
	\end{center}
	\vspace{-0.4cm}
\end{figure}

We tested whether the segmentation accuracy of the motion elements could be improved by mutual learning of the two layers.
The sequence of correct labels for the evaluation was established based on the workflow presented in Table \ref{tbl:class}.
Table \ref{fig:ham} shows the NLDs between the segmented motion elements and correct labels.
The three methods with hierarchical mutual learning had a smaller NLD than GP-HSMM alone. 
This result indicates that the mutual-learning method, which uses the probability of generating motion elements from unit motion as a prior distribution for training the GP-HSMM, works effectively.

Histograms of the NLDs obtained from 10 trials for each method are shown in Figure \ref{fig:ham}.
The histograms of the three proposed methods with mutual learning are more biased to the left than those of the GP-HSMM alone without mutual learning.
This indicates that mutual learning tends to improve the segmentation accuracy of motion elements.

\begin{table}[t]
  \caption{Normalized Levenshtein distance between the estimated segment and the ground truth. (WS: word segmentation, ME-U: motion element unigram, MEB: motion element bigram)}
  \label{table:hd}
  \centering
  \begin{tabular}{ccc}
    \hline
     Method  & NLD   \\
    \hline \hline
    GP-HSMM & 0.59\\
    GP-HSMM + WS HSMM &  0.50  \\
    GP-HSMM + ME-U HSMM  & 0.50   \\
    GP-HSMM + ME-B HSMM  &  0.52   \\
    \hline
  \end{tabular}
\end{table}

\subsection{Segmentation of unit motions}

\begin{table*}[t]
\caption{Task procedures}
\label{table:Unit m}
\centering
\begin{tabular}{ccll}
 \hline
 Motion Label& Unit Motion Label & Task Description & Unit motion  \\
 \hline\hline
1 & &Take part A from the cart and &\\ 
   & 1 & ~~~~place it on the workspace & Fasten parts A and B\\
2 & & Take part B from the box &\\
3 & & Attach part B to part A  &\\
\hline
4 & & Take a screw from the box  &\\
5 & & Fix part B and part A with the screw &\\
4 & 2 &Take a screw from the box&  Fix using a screw and a screw driver\\
5 & & Fix part B and part A with the screw &\\
4 & & Take a screw from the box &\\
5 & & Fix part B and part A with the screw &\\
\hline
6 & & Take part C from the box &\\
   & & ~~~~and attach it to part B &  \\
7 & 3 & Take part D from the cart & Fasten parts C and D\\
8 & & Attach part D to part A and &\\
   & & ~~~~place the finished product on the cart &\\
\hline
\end{tabular}
\end{table*}

Subsequently, we compared the accuracy of the unit motions estimated using the three methods.  
Three unit operations (Fasten parts A and B, Fix using a screw and screwdriver, and Fasten parts C and D) were used as correct answers, as shown in Table \ref{table:Unit m}.
The results are presented in Table \ref{table:lei}.
The word-segmentation HSMM had the highest value, and the difference from the correct labels was large.
This is because the word-segmentation HSMM classified all the different sequences of motion elements into different unit motions, and could not absorb the fluctuations of the actions and procedures.
By contrast, the two models that used the generation probability of the motion elements could classify slightly different motion elements into the same unit motion class.

The number of unit motions estimated by each method was 140 for the word segmentation HSMM, 6 for the motion element unigram HSMM, and 6 for the motion element bigram HSMM.
The word segmentation HSMM significantly increases the number of unit motions by classifying all patterns of sequences of motion elements caused by fluctuations in behaviors and procedures into different unit motions.

These results indicate that it is difficult to properly segment noisy data with a word-segmentation HSMM using simple word segmentation.
However, the two models that use the probability of generating motion elements can segment noisy real data with higher accuracy.

\begin{figure}[t]
	\begin{center}
	\includegraphics[scale=0.65]{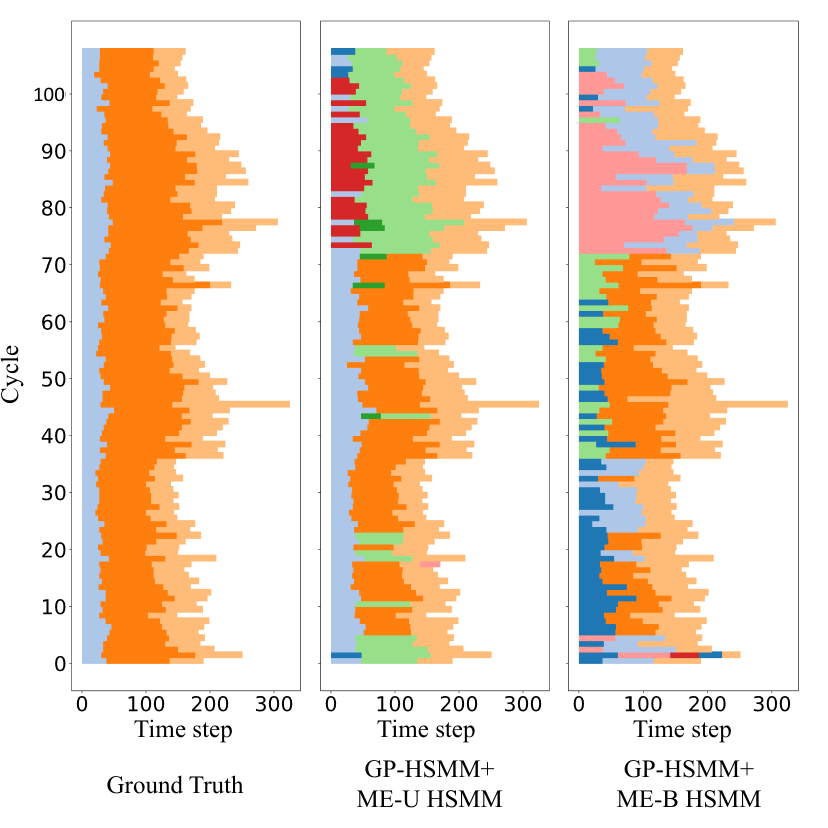}
	\caption{Visualization of unit motion segmentation. Left: ground truth, Middle: segments estimated by GP-HSMM+ME-U HSMM, and Right: segments estimated by GP-HSMM+ME-B HSMM. }
	\label{fig:um results}
	\end{center}
	\vspace{-0.4cm}
\end{figure}

Figure \ref{fig:um results} shows a graph that visualizes the segmentation of the unit motion of the ground truth, the segmentation estimated by GP-HSMM+ME-U HSMM, and the segmentation estimated by GP-HSMM+ME-B HSMM. 
The horizontal axis represents the time step, and the vertical axis represents the operation (1-36 represents the cycle of the 1st worker, 37-72 represents the 2nd worker, and 73-108 represents the 3rd worker). The same color indicates the same class of indices classified by each HSMM.
Figure \ref{fig:um results} shows that the segmentation of the motion element unigram is more similar to the ground truth. 
Therefore, the motion element unigram HSMM fits these data better than the motion element bigram.

\begin{table}[t]
  \caption{ Levenshtein distance between the estimated segment and the ground truth}
  \label{table:lei}
  \centering
  \begin{tabular}{ccc}    \hline
     Method & NLD   \\ \hline \hline
    GP-HSMM + WS HSMM & 0. 90    \\
    GP-HSMM + ME-U HSMM  & 0.33  \\
    GP-HSMM + ME-B HSMM  &   0.38  \\
    \hline
  \end{tabular}
\end{table}

%

\section{Conclusion} 
In this study, we proposed a PGM that performs mutual learning in two layers using unsupervised learning, which does not require labeled data.
Furthermore, we compared three models with different emission probabilities for the upper layer, HSMM.
In the experiment, we used the coordinates of both wrists of three workers performing cell production on a shop floor, and revealed that segmentation can be performed more accurately than in the conventional method by mutual learning in two layers.
Furthermore, we demonstrated that the HSMM with a motion element unigram as the emission is the most effective method for real data, such as the data used in this experiment, where there is a variation in behaviors.
We believe that these emission probabilities should be selected appropriately depending on the nature of the data used.
For example, for noiseless data, it is effective to use the word-segmentation model to distinguish small differences in the behavior clearly.
However, if the same operation can be conducted using different motion elements, such as for the data used in this study, a unigram model is effective.
If the same motion elements occur frequently in different unit tasks, the order of the motion elements is important in distinguishing them. 
In this case, the unit-motion bigram HSMM model is considered effective.
In the future, we plan to clarify the relationship between the nature of the data and emission probability.

In addition to this issue, the current method has some limitations. 
The first limitation is that the GP-HSMM and the HSMM require knowledge regarding the number of classes in advance. 
For GP-HSMM, this limitation can be addressed using a nonparametric Bayesian model HDP-GP-HSMM in which a hierarchical Dirichlet process (HDP) is introduced into the GP-HSMM \cite{nagano2018sequence}. 
Similarly, we believe that it is possible to estimate the number of classes in HSMM. 
The second limitation is computational cost, particularly in the GP-HSMM. The computational cost of training a Gaussian process is $O(n^3)$, where $n$ represents the length of a sequence. 
We regard solving this problem as essential to applying our method to larger data. 
We believe that this issue can be resolved by introducing Gaussian processes with lower computational costs \cite{nguyen2008local, okadome2014adaptive, gardner2018gpytorch}. 
Furthermore, to reduce the computational cost, we can explore the possibility of bypassing computations at the points that are less likely boundaries by employing slice sampling \cite{10.1214/aos/1056562461} to truncate the forward probability. 

Manual behavior analysis by experts watching videos, a current primary method, is time-consuming. 
Our proposed solution automates the segmentation process, facilitating automatic behavior analysis, thereby potentially enabling feedback to be provided swiftly without experts. 
However, to effectively apply the current methodology in real-world scenarios, a method for effectively visualizing analysis results and an easy-to-use application need to be developed. 
Additionally, as the current computation is offline, addressing the real-time computation is also part of future work.
\bibliographystyle{IEEEtran}
\bibliography{reference}


\begin{IEEEbiography}[{\includegraphics[width=1in,height=1.25in,clip,keepaspectratio]{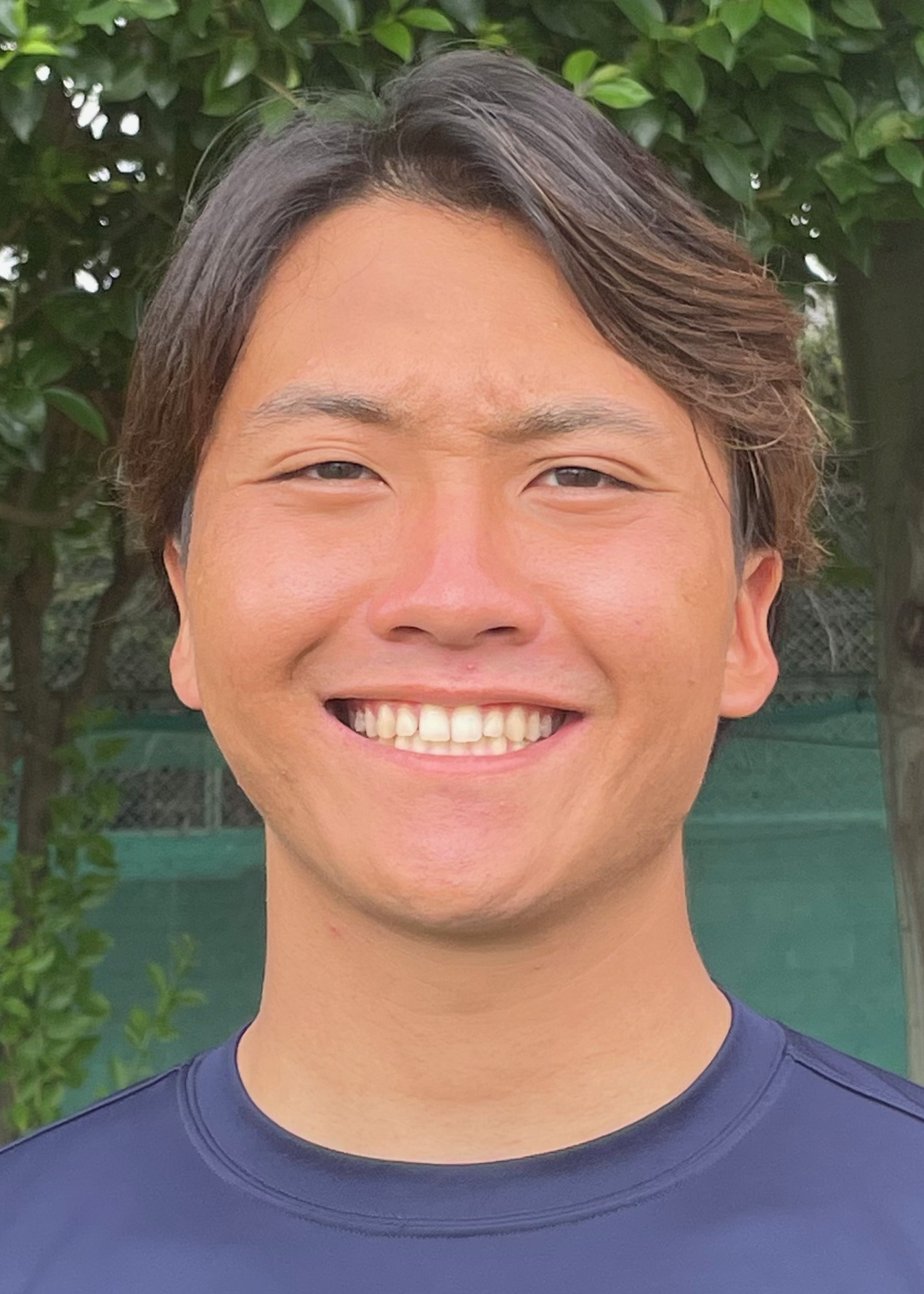}}]{Issei Saito} 
Issei Saito received his Bachelor's degree from the University of Electro-Communications in 2023. He is currently pursuing a Master's degree at the Graduate School of Informatics and Engineering, the University of Electro-Communications. His research interests include intelligent robotics and machine learning.
\end{IEEEbiography}

\begin{IEEEbiography}[{\includegraphics[width=1in,height=1.25in,clip,keepaspectratio]{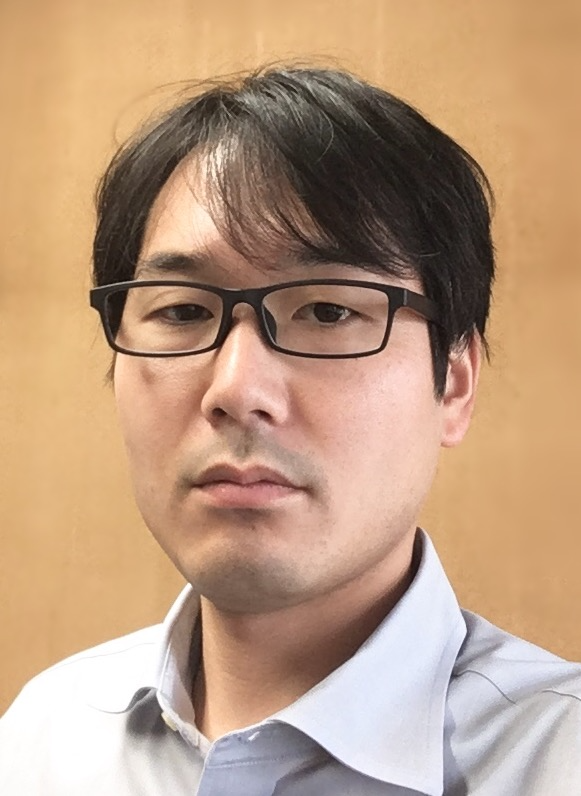}}]{Tomoaki Nakamura} 
received his BE, ME, and Dr. of Eng. degrees from the
University of Electro-Communications in 2007, 2009, and 2011.
From April 2011 to March 2012, He was a research fellow of the Japan Society for
the Promotion of Science.
In 2013, he worked for Honda Research Institute Japan Co., Ltd. 
From April 2014 to March 2018, he was an Assistant Professor at the Department of Mechanical Engineering and Intelligent Systems, the University of Electro-Communications. 
Since April 2019, he has been an Associate Professor at the same department.
His research interests include intelligent robotics and machine learning.
He has received the IROS Best Paper Award Finalist, the Advanced Robotics Best Paper Award, and the JSAI Best Paper Award.
\end{IEEEbiography}

\begin{IEEEbiography}[{\includegraphics[width=1in,height=1.25in,clip,keepaspectratio]{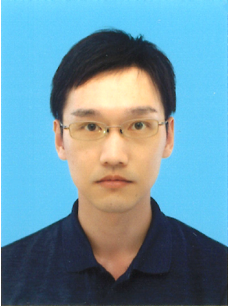}}]{Toshiyuki Hatta}
received his ME from the Graduate School of Engineering Science, Osaka University in 2014. He is currently a senior researcher at the Advanced Technology R\&D Center, Mitsubishi Electric Corp., and a Ph.D. student at the Graduate School of Informatics and Engineering, the University of Electro-Communications. His research interests include machine learning and computer vision.
\end{IEEEbiography}
\begin{IEEEbiography}[{\includegraphics[width=1in,height=1.25in,clip,keepaspectratio]{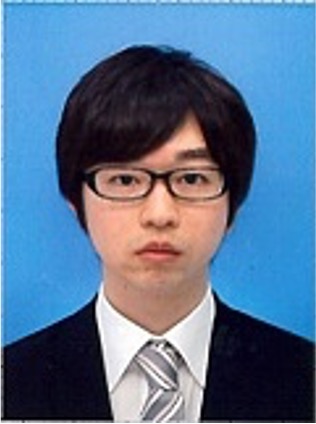}}]{Wataru Fujita}
received an ME from the Graduate School of Information Science and Technology, Osaka University. He is a researcher at the Advanced Technology R\&D center, Mitsubishi Electric Corp. His research interests include machine learning and computer vision.
\end{IEEEbiography}
\begin{IEEEbiography}[{\includegraphics[width=1in,height=1.25in,clip,keepaspectratio]{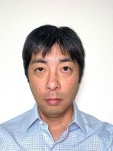}}]{Shintaro Watanabe}
received an ME from Kyoto University, Japan. He is currently a senior manager at the Advanced Technology R\&D Center, Mitsubishi Electric Corp. His research mainly includes image recognition, computer vision, machine learning, and deep learning. He leads image recognition projects for industrial applications.
\end{IEEEbiography}
\begin{IEEEbiography}[{\includegraphics[width=1in,height=1.25in,clip,keepaspectratio]{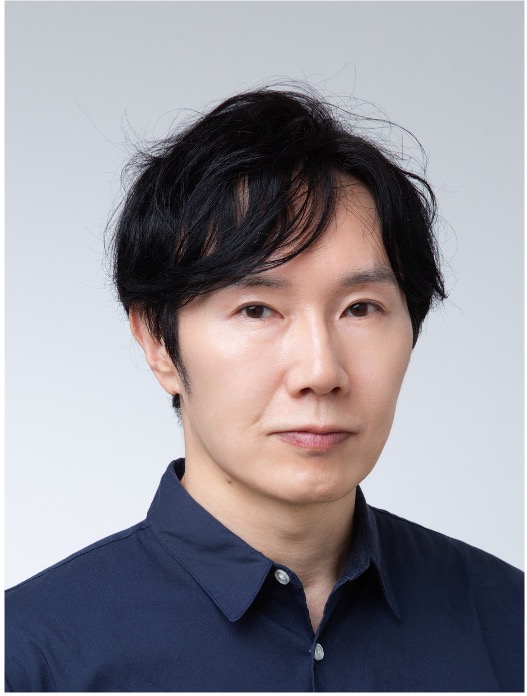}}]{Shotaro Miwa}
received a BE from the University of Tokyo, Japan, and a PhD degree from Osaka University, Japan. He is currently a chief researcher at the Information Technology R\&D Center, Mitsubishi Electric Corp., a researcher at the National Institute of Advanced Industrial Science and Technology (AIST), and a visiting researcher at the University of Alberta. His research mainly includes machine learning, computer vision, deep learning, and deep reinforcement learning. He leads artificial intelligence projects for industrial applications.
\end{IEEEbiography}

\EOD

\end{document}